\documentclass[10pt,twocolumn,letterpaper]{article}

\usepackage{iccv}
\usepackage{times}
\usepackage{epsfig}
\usepackage{graphicx}
\usepackage{amsmath}
\usepackage{amssymb}
\usepackage{mathtools}
\usepackage{multirow}
\usepackage{pifont}
\usepackage{subfigure} 
\usepackage{adjustbox} 

\newcommand\numberthis{\addtocounter{equation}{1}\tag{\theequation}}

\newcommand{\Umat}{{\bf U}}
\newcommand{\Vmat}{{\bf V}}
\newcommand{\Wmat}{{\bf W}}

\newcommand{\bv}{{\boldsymbol b}}
\newcommand{\cv}{{\boldsymbol c}}

\newcommand{\qv}{{\boldsymbol q}}

\newcommand{\vv}{{\boldsymbol v}}
\newcommand{\wv}{{\boldsymbol w}}

\newcommand{\R}{\mathbb{R}}

\newcommand{\Acal}{\mathcal{A}}

\newcommand{\Ncal}{\mathcal{N}}

\newcommand{\Gcal}{\mathcal{G}}

\newcommand{\Vcal}{\mathcal{V}}

\newcommand{\Ecal}{\mathcal{E}}

\usepackage[pagebackref=true,breaklinks=true,letterpaper=true,colorlinks,bookmarks=false]{hyperref}

\iccvfinalcopy 

\def\iccvPaperID{3154} 

\ificcvfinal\fi
\begin{document}

\title{Relation-Aware Graph Attention Network for Visual Question Answering}

\author{Linjie Li, Zhe Gan, Yu Cheng, Jingjing Liu  \\
Microsoft Dynamics 365 AI Research  \\
{\tt\small \{lindsey.li, zhe.gan, yu.cheng, jingjl\}@microsoft.com}
}

\maketitle

\begin{abstract}
In order to answer semantically-complicated questions about an image, a Visual Question Answering (VQA) model needs to fully understand the visual scene in the image, especially the interactive dynamics between different objects. We propose a Relation-aware Graph Attention Network (ReGAT), which encodes each image into a graph and models multi-type inter-object relations via a graph attention mechanism, to learn question-adaptive relation representations. Two types of visual object relations are explored: ($i$) Explicit Relations that represent geometric positions and semantic interactions between objects; and ($ii$) Implicit Relations that capture the hidden dynamics between image regions. Experiments demonstrate that ReGAT outperforms prior state-of-the-art approaches on both VQA 2.0 and VQA-CP v2 datasets. We further show that ReGAT is compatible to existing VQA architectures, and can be used as a generic relation encoder to boost the model performance for VQA.\footnote{Code is available at \url{https://github.com/linjieli222/VQA_ReGAT}.}
\end{abstract}

\vspace{-4mm}
\section{Introduction}
Recent advances in deep learning have driven tremendous progress in both Computer Vision and Natural Language Processing (NLP). Interdisciplinary area between language and vision, such as image captioning, text-to-image synthesis and visual question answering (VQA), has attracted rapidly growing attention from both vision and NLP communities. Take VQA as an example - the goal (and the main challenge) is to train a model that can achieve comprehensive and semantically-aligned understanding of multimodal input. Specifically, given an image and a natural language question grounded on the image, the task is to associate visual features in the image with the semantic meaning in the question, in order to correctly answer the question.

Most state-of-the-art approaches to VQA \cite{yang2016stacked,fan2018stacked,nam2016dual,lu2016hierarchical,teney2017tips} focus on learning a multimodal joint representation of images and questions. Specifically, a Convolutional Neural Network (CNN) or Region-based CNN (R-CNN) is commonly used as a visual feature extractor for image encoding, and a Recurrent Neural Network (RNN) is used for question encoding. After obtaining a sparse set of image regions from the visual feature extractor, multimodal fusion is applied to learn a joint representation that represents the alignment between each individual region and the question.
This joint representation is then fed into an answer predictor to produce an answer. 

\begin{figure}[t]
 \centering
 \includegraphics[width = 0.48\textwidth]{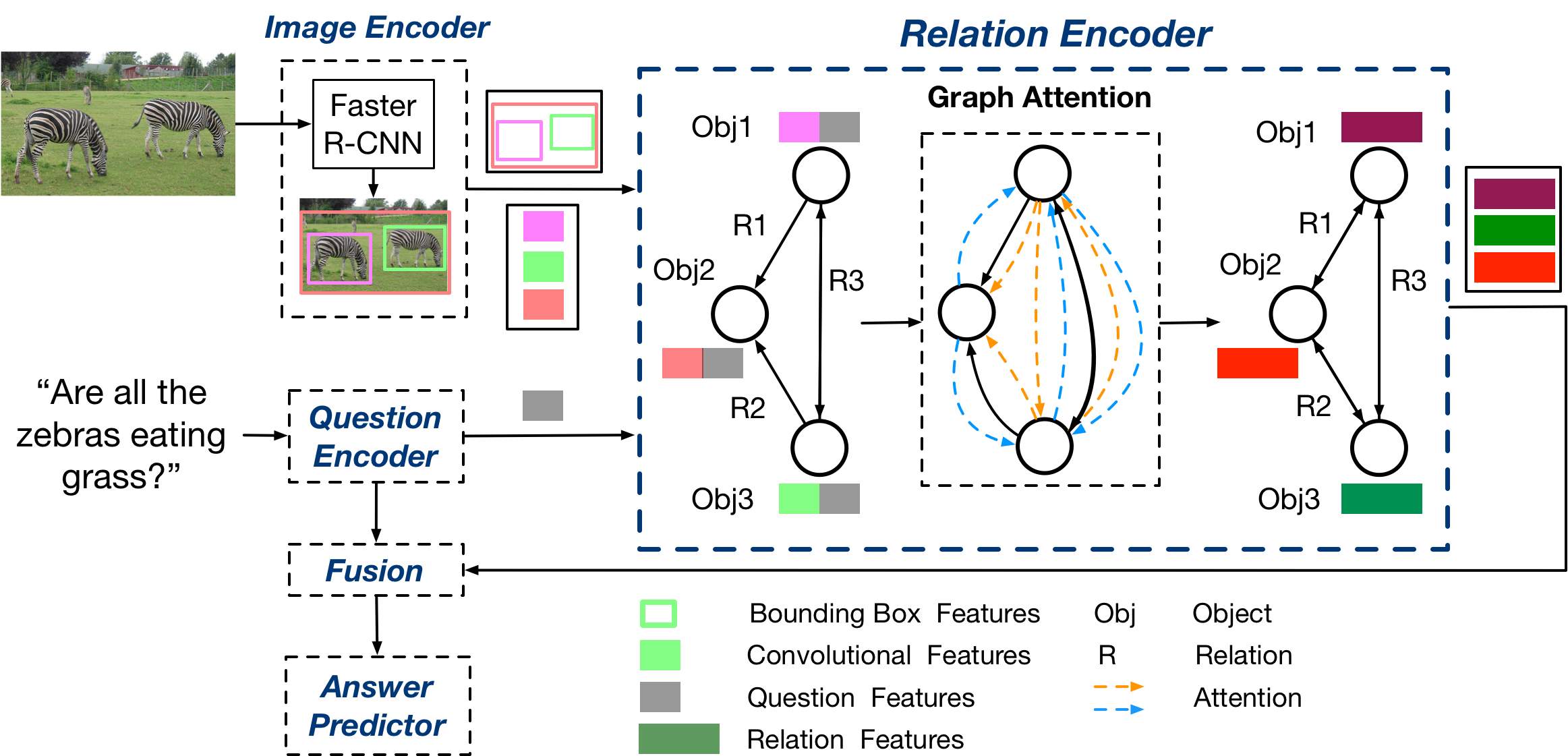}
 \caption{\small{An overview of the ReGAT model. Both explicit relations (semantic and spatial) and implicit relations are considered. The proposed relation encoder captures question-adaptive object interactions via Graph Attention.}}
 \label{fig:VQA_arch}
\vspace{-4mm}
\end{figure}

This framework has proven to be useful for the VQA task, but there still exists a significant semantic gap between image and natural language. For example, given an image of a group of zebras (see Figure~\ref{fig:VQA_arch}), the model may recognize the black and white pixels, but not which white and black pixels are from which zebra. Thus, it is difficult to answer questions such as ``\emph{Is the zebra at the far right a baby zebra?}" or ``\emph{Are all the zebras eating grass?}". A VQA system needs to recognize not only the objects (``\emph{zebras}``) and the surrounding environment (``\emph{grass}''), but also the semantics about actions (``\emph{eating}'') and locations (``\emph{at the far right}'') in both images and questions.

In order to capture this type of action and location information, we need to go beyond mere object detection in image understanding, and learn a more holistic view of the visual scene in the image, by interpreting the dynamics and interactions between different objects in an image. One possible solution is to detect the relative geometrical positions of neighboring objects (e.g., \textless \texttt{motorcycle-next to-car}\textgreater), to align with spacial descriptions in the question. Another direction is to learn semantic dependencies between objects (e.g., \textless \texttt{girl-eating-cake}\textgreater) to capture the interactive dynamics in the visual scene.

Motivated by this, we propose a Relation-aware Graph Attention Network (ReGAT) for VQA, introducing a novel relation encoder that captures these inter-object relations beyond static object/region detection. These visual relation features can reveal more fine-grained visual concepts in the image, which in turn provides a holistic scene interpretation that can be used for answering semantically-complicated questions. In order to cover the high variance in image scenes and question types, both explicit (e.g., spatial/positional, semantic/actionable) relations and implicit relations are learned by the relation encoder, where images are represented as graphs and interactions between objects are captured via a graph attention mechanism.

Furthermore, the graph attention is learned based on the context of the question, permitting the injection of semantic information from the question into the relation encoding stage. In this way, the features learned by the relation encoder not only capture object-interactive visual contents in the image, but also absorb the semantic clues in the question, to dynamically focus on particular relation types and instances for each question on the fly.    

Figure~\ref{fig:VQA_arch} shows an overview of the proposed model. First, a Faster R-CNN is used to generate a set of object region proposals, and a question encoder is used for question embedding. The convolutional and bounding-box features of each region are then injected into the relation encoder to learn the relation-aware, question-adaptive, region-level representations from the image.
These relation-aware visual features and the question embeddings are then fed into a multimodal fusion module to produce a joint representation, which is used in the answer prediction module to generate an answer. 

In principle, our work is different from (and compatible to) existing VQA systems. It is pivoted on a new dimension: using question-adaptive inter-object relations to enrich image representations in order to enhance VQA performance.
The contributions of our work are three-fold:
\begin{itemize}
    \vspace{-2mm}
    \item We propose a novel graph-based relation encoder to learn both explicit and implicit relations between visual objects via graph attention networks. 
    \vspace{-2mm}
    \item The learned relations are question-adaptive, meaning that they can dynamically capture visual object relations that are most relevant to each question. 
    \vspace{-2mm}
    \item We show that our ReGAT model is a generic approach that can be used to improve state-of-the-art VQA models on the VQA 2.0 dataset. Our model also achieved state-of-the-art performance on the more challanging VQA-CP v2 dataset. 
\end{itemize}

\section{Related Work} \label{related_work}
\subsection{Visual Question Answering} 

The current dominant framework for VQA systems 
consists of an image encoder, a question encoder, multimodal fusion, and an answer predictor. 
In lieu of directly using visual features from CNN-based feature extractors, \cite{yang2016stacked,fan2018stacked,patro2018differential,lu2016hierarchical,teney2017tips,nam2016dual,zhu2017structured,malinowski2018learning} explored various image attention mechanisms to locate regions that are relevant to the question. To learn a better representation of the question, \cite{lu2016hierarchical,nam2016dual,fan2018stacked} proposed to perform question-guided image attention and image-guided question attention collaboratively, to merge knowledge from both visual and textual modalities in the encoding stage. \cite{fukui2016multimodal,kim2016hadamard,yu2018beyond,ben2017mutan,kim18bilinear} explored higher-order fusion methods to better combine textual information with visual information (e.g., using bilinear pooling instead of simpler first-order methods such as summation, concatenation and multiplication).

To make the model more interpretable, some literature \cite{li2018tell,yu2017multi, li2017incorporating,wu2017image,wu2016ask,wang2017vqa} also exploited high-level semantic information in the image, such as attributes, captions and visual relation facts. Most of these methods applied VQA-independent models to extract semantic knowledge from the image, while \cite{lu2018r} built a Relation-VQA dataset and directly mined VQA-specific relation facts to feed additional semantic information to the model. A few recent studies \cite{su2018learning,ma2018visual,li2017incorporating} investigated how to incorporate memory to aid the reasoning step, especially for difficult questions. 

However, the semantic knowledge brought in by either memory or high-level semantic information is usually converted into textual representation, instead of directly used as visual representation, which contains richer and more indicative information about the image. Our work is complementary to these prior studies in that we encode object relations directly into image representation, and the relation encoding step is generic and can be naturally fit into any state-of-the-art VQA model.

\begin{figure*}[ht]
 \centering
 \includegraphics[width =\textwidth]{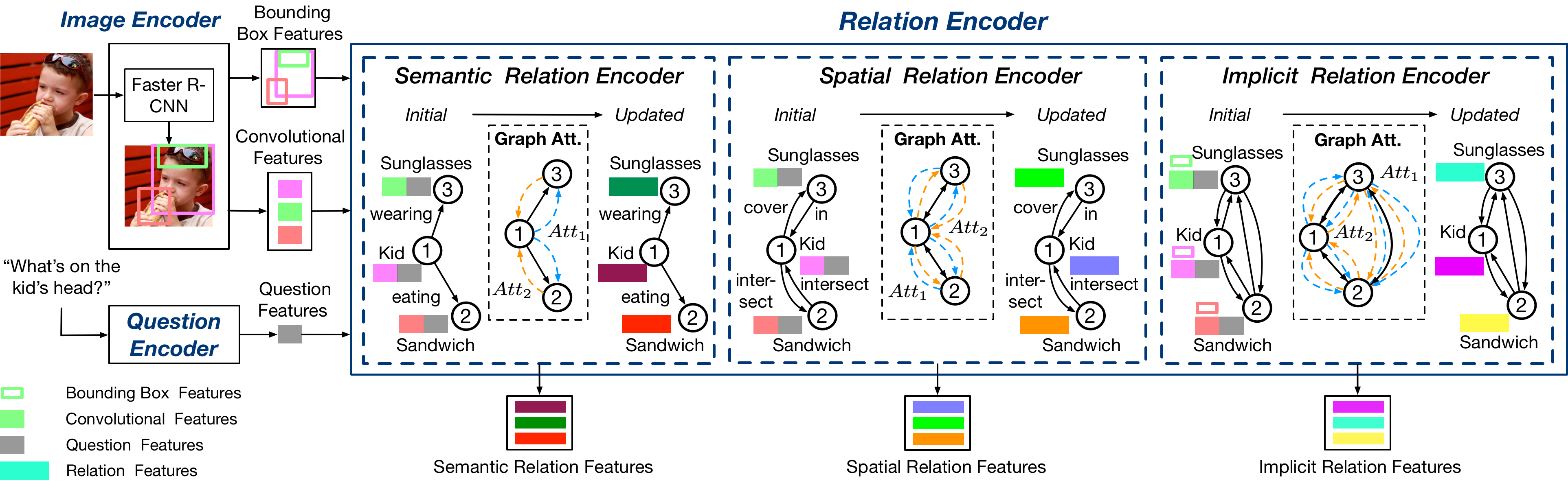}
 \caption{\small{Model architecture of the proposed ReGAT for visual question answering. Faster R-CNN is employed to detect a set of object regions. These region-level features are then fed into different relation encoders to learn relation-aware question-adaptive visual features, which will be fused with question representation to predict an answer. Multimodal fusion and answer predictor are omitted for simplicity.}}
 \vspace{-4mm}
 \label{fig:our_model}
\end{figure*}

\subsection{Visual Relationship} \label{sec:visual_relationship} 
Visual relationship has been explored before deep learning became popular. Early work \cite{divvala2009empirical,felzenszwalb2010object,choi2012tree,mottaghi2014role} presented methods to re-score the detected objects by considering object relations (e.g.,  co-occurrence~\cite{divvala2009empirical}, position and size~\cite{biederman1982scene}) as post-processing steps for object detection. Some previous work \cite{galleguillos2008object,gould2008multi} also probed the idea that spatial relationships (e.g., ``above", ``around", ``below" and ``inside") between objects can help improve image segmentation. 

Visual relationship has proven to be crucial to many computer vision tasks.
For example, it aided the cognitive task of mapping images to captions \cite{farhadi2010every,fang2015captions,yao2018exploring} and improved image search \cite{schuster2015generating,johnson2015image} and object localization \cite{sadeghi2011recognition,hu2018relation}. Recent work on visual relationship \cite{sadeghi2011recognition,ramanathan2015learning,divvala2014learning} focused more on non-spatial relation, or known as ``semantic relation'' (i.e., actions of, or interactions between objects). 
A few neural network architectures have been designed for the visual relationship prediction task \cite{lu2016visual,dai2017detecting,zhang2017visual}.  

\subsection{Relational Reasoning}
We name the visual relationship aforementioned as \emph{explicit} relation, which has been shown to be effective for image captioning~\cite{yao2018exploring}. Specifically, \cite{yao2018exploring} exploited pre-defined semantic relations learned from the Visual Genome dataset \cite{krishna2017visual} and spatial relations between objects. A graph was then constructed based on these relations, and a Graph Convolutional Network (GCN)~\cite{kipf2016semi} was used to learn representations for each object.

Another line of research focuses on \emph{implicit} relations, where no explicit semantic or spatial relations are used to construct the graph. Instead, all the relations are \emph{implicitly} captured by an attention module or via higher-order methods over the fully-connected graph of an input image~\cite{santoro2017simple,hu2018relation,cadene2019murel,yang2018multi}, to model the interactions between detected objects. For example,~\cite{santoro2017simple} reasons over
all the possible pairs of objects in an image via the use of simple MLPs. In~\cite{cadene2019murel}, a bilinear fusion method, called MuRel cell, was introduced to perform pairwise relationship modeling.

Some other work~\cite{teney2017graph,norcliffe2018learning,wang2019neighbourhood} have been proposed for learning question-conditioned graph representations for images. Specifically, \cite{norcliffe2018learning} introduced a graph learner module that is conditioned on question representations to compute the image representations using pairwise attention and spatial graph convolutions. \cite{teney2017graph} exploited structured question representations such as parse trees, and used GRU to model contextualized interactions between both objects and words. A more recent work \cite{wang2019neighbourhood} introduced a sparser graph defined by inter/intra-class edges, in which relationships are implicitly learned via a language-guided graph attention mechanism. However, all these work still focused on \emph{implicit} relations, which are less interpretable than \emph{explicit} relations.

\paragraph{Our contributions}
Our work is inspired by~\cite{hu2018relation,yao2018exploring}. However, different from them, ReGAT considers both explicit and implicit relations to enrich image representations. For explicit relations, our model uses Graph Attention Network (GAT) rather than a simple GCN as used in~\cite{yao2018exploring}. As opposed to GCNs, the use of GAT allows for assigning different importances to nodes of the same neighborhood. For implicit relations, our model learns a graph that is adaptive to each question by filtering out question-irrelevant relations, instead of treating all the relations equally as in~\cite{hu2018relation}. In experiments, we conduct detailed ablation studies to demonstrate the effectiveness of each individual design.

\section{Relation-aware Graph Attention Network}
Here is the problem definition of the VQA task: given a question $q$ grounded in an image $I$, the goal is to predict an answer $\hat{a} \in \Acal$ that best matches the ground-truth answer $a^\star$. As common practice in the VQA literature, this can be defined as a classification problem:
\begin{align}
    \hat{a} = \arg \max_{a\in \Acal} p_{\theta} (a|I,q)\,,
\end{align}
where $p_{\theta}$ is the trained model. 

Figure~\ref{fig:our_model} gives a detailed illustration of our proposed model, consisting of an Image Encoder, a Question Encoder, and a Relation Encoder. 
For the Image Encoder, Faster R-CNN~\cite{anderson2018bottom} is used to identify a set of objects $\Vcal=\{v_i\}_{i=1}^K$, where each object $v_i$ is associated with a visual feature vector $\vv_i \in \R^{d_v}$ and a bounding-box feature vector $\bv_i \in \R^{d_b}$ ($K=36$, $d_v=2048$, and $d_b=4$ in our experiments). Each $\bv_i=[x,y,w,h]$ corresponds to a 4-dimensional spatial coordinate, where $(x,y)$ denotes the coordinate of the top-left point of the bounding box, and $h$/$w$ corresponds to the height/width of the box. For the Question Encoder, we use a bidirectional RNN with Gated Recurrent Unit (GRU) and perform self attention on the sequence of RNN hidden states to generate question embedding $\qv\in \R^{d_q}$ ($d_q=1024$ in our experiments). The following sub-sections will explain the details of the Relation Encoder.

\subsection{Graph Construction}
\paragraph{Fully-connected Relation Graph}
By treating each object $v_i$ in the image as one vertex, we can construct a fully-connected undirected graph $\Gcal_{imp} = (\Vcal,\Ecal)$, where $\Ecal$ is the set of $K\times (K-1)$ edges. Each edge represents an implicit relation between two objects, which can be reflected by the learned weight assigned to each edge through graph attention. All the weights are learned implicitly without any prior knowledge.
We name the relation encoder built on this graph $\Gcal_{imp}$ the \emph{implicit} relation encoder. 

\vspace{-2mm}
\paragraph{Pruned Graph with Prior Knowledge}
On the other hand, if explicit relations between vertices are available, one can readily transform the fully-connected graph $\Gcal_{imp}$ into an explicit relation graph, by pruning the edges where the corresponding explicit relation does not exist. For each pair of objects $i,j$, if \textless$i$-$p$-$j$\textgreater is a valid relation, an edge is created from $i$ to $j$, with an edge label $p$. In addition, we assign each object node $i$ with a self-loop edge and label this edge as \texttt{identical}.
In this way, the graph becomes sparse, and each edge encodes prior knowledge about one inter-object relation in the image. We name the relation encoder built upon this graph the \emph{explicit} relation encoder. 

The explicit nature of these features requires pre-trained classifiers to extract the relations in the form of discrete class labels, which represent the dynamics and interactions between objects explicit to the human eye. Different types of explicit relations can be learned based on this pruned graph. In this paper, we explore two instances: spatial and semantic graphs, to capture positional and actionable relations between objects, which is imperative for visual question answering.

\begin{figure}[t!]
    \centering
    \subfigure[\small{Spatial Relation}]{
		\includegraphics[width=.16\textwidth]{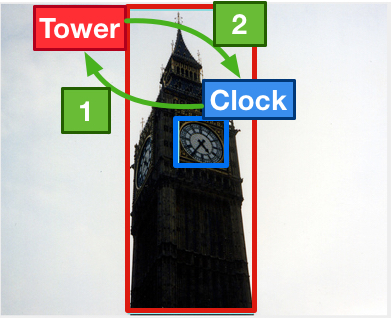}
		\label{fig:spatial_relation}
}%
\qquad
\subfigure[\small{Semantic Relation}]{
		\includegraphics[width=.20\textwidth]{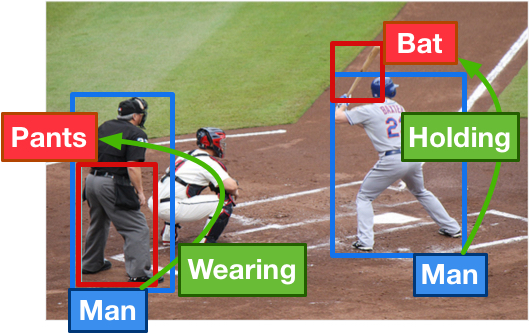}
		\label{fig:semantic_relation}
}
    \caption{\small{Illustration of spatial and semantic relations. The green arrows denote the direction of relations (subject $\rightarrow$  object). Labels in green boxes are class labels of relations. Red and Blue boxes contain class labels of objects.}}
    \label{fig:relation_example}
\vspace{-4mm}
\end{figure}

\textbf{Spatial Graph} Let $spa_{i,j} = $\textless \texttt{$\texttt{object}_i$-\texttt{predicate} -$\texttt{object}_j$}\textgreater \space denote the spatial relation that represents the relative geometric position of $\texttt{object}_i$ against $\texttt{object}_i$.
In order to construct a spatial graph $\Gcal_{spa}$,
given two object region proposals $\texttt{object}_i$ and $\texttt{object}_j$, 
we classify $spa_{i,j}$ into $11$ different categories~\cite{yao2018exploring} (e.g., $\texttt{object}_i$ is inside $\texttt{object}_j$ (class 1), $\texttt{object}_j$ is inside $\texttt{object}_i$ (class 2), as illustrated in Figure~\ref{fig:spatial_relation}), including a \texttt{no-relation} class retained for objects that are too far away from each other.
Note that edges formed by spatial relations are symmetrical: if \textless \texttt{$\texttt{object}_i$-$p_{i,j}$-$\texttt{object}_j$} \textgreater \space is a valid spatial relation, there must be a valid spatial relation  $spa_{j,i} = $\textless \texttt{$\texttt{object}_j$-$p_{j,i}$-$\texttt{object}_i$} \textgreater. However, the two predicates $p_{i,j}$ and $p_{j,i}$ are different. 

\textbf{Semantic Graph} In order to construct semantic graph $\Gcal_{sem}$, semantic relations between objects need to be extracted (e.g., \textless \texttt{subject-predicate-object}\textgreater). This can be formulated as a classification task \cite{yao2018exploring} by training a semantic relation classifier on a visual relationship dataset (e.g., Visual Genome \cite{krishnavisualgenome}). Given two object regions $i$ and $j$, the goal is to determine which predicate $p$ represents a semantic relation \textless \texttt{$i$-$p$-$j$}\textgreater \space between these two regions. Here, the relations between the subject $j$ and the object $i$ are not interchangeable, meaning that the edges formed by semantic relations are not symmetric. For a valid \textless \texttt{$i$-$p_{i,j}$-$j$}\textgreater, there may not exist a relation \textless \texttt{$j$-$p_{j.i}$-$i$}\textgreater \space within our definition. For example,  \textless \texttt{man-holding-bat}\textgreater \space is a valid relation, while  there is no semantic relation from \texttt{bat} to \texttt{man}. 

The classification model takes in three inputs: feature vector of the subject region $\vv_i$, feature vector of the object region $\vv_j$, and region-level feature vector $\vv_{i,j}$ of the union bounding box containing both $i$ and $j$. These three types of feature are obtained from pre-trained object detection model, and then transformed via an embedding layer. The embedded features are then concatenated and fed into a classification layer to produce softmax probability over 14 semantic relations, with an additional \texttt{no-relation} class. 
The trained classifier is then used to predict relations between any pair of object regions in a given image. Examples of semantic relations are shown in Figure~\ref{fig:semantic_relation}.

\subsection{Relation Encoder} \label{sec:imp_relation}

\paragraph{Question-adaptive Graph Attention}
The proposed relation encoder is designed to encode relational dynamics between objects in an image. For the VQA task, there might be different types of relations that are useful for different question types. Thus, in designing the relation encoder, we use a question-adaptive attention mechanism to inject semantic information from the question into relation graphs, to dynamically assign higher weights to those relations that are mostly relevant to each question. 
This is achieved by first concatenating the question embedding $\qv$ with each of the $K$ visual features $\vv_i$, denoted as
\begin{align}
    \vv_i^{\prime} = [\vv_i||\qv] \quad \mbox{for} \,\, i=1,\ldots,K\,.
\end{align}
Self-attention is then performed on the vertices, which generates hidden relation features $\{\vv_i^{\star}\}_{i=1}^K$ that characterize the relations between a target object and its neighboring objects. 
Based on this, each relation graph goes through the following attention mechanism: 
\begin{align}\label{ref:hid_rel_eq_1}
    \vv_i^{\star} = \sigma\Big(\sum_{j\in \Ncal_i} \alpha_{ij}\cdot \Wmat \vv_j^{\prime} \Big)\,.
\end{align}
For different types of relation graph, the definition of the attention coefficients $\alpha_{ij}$ varies, so does the projection matrix $\Wmat\in \R^{d_h \times (d_q+d_v)}$ and the neighborhood $\Ncal_i$ of object $i$. $\sigma(\cdot)$ is a nonlinear function such as ReLU. To stabilize the learning process of self-attention, we also extend the above graph attention mechanism by employing multi-head attention, where $M$ independent attention mechanisms are executed, and their output features are concatenated, resulting in the following output feature representation:
\begin{align}\label{ref:hid_rel_eq_2}
    \vv_i^{\star} =\Vert_{m=1}^{M} \sigma\Big(\sum_{j\in \Ncal_i} \alpha_{ij}^m\cdot \Wmat^m \vv_j^{\prime} \Big)\,.
\end{align}
In the end, $\vv_i^{\star}$ is added to the original visual feature $\vv_i$ to serve as the final relation-aware feature.

\paragraph{Implicit Relation}
Since the graph for learning implicit relation is fully-connected, $\Ncal_i$ contains all the objects in the image, including object $i$ itself.
Inspired by~\cite{hu2018relation}, we design the attention weight $\alpha_{ij}$ to not only depend on visual-feature weight $\alpha^{v}_{ij}$, but also bounding-box weight $\alpha^{b}_{ij}$. Specifically,
\begin{align}\label{ref:hid_rel_eq_2}
    \alpha_{ij}&= \frac{\alpha^{b}_{ij} \cdot \exp(\alpha^{v}_{ij})}{\sum_{j=1}^K \alpha^{b}_{ij} \cdot \exp(\alpha^{v}_{ij})} \,,
\end{align}
where $\alpha^{v}_{ij}$ represents the similarity between the visual features, computed by scaled dot-product~\cite{vaswani2017attention}:
\begin{align}\label{ref:hid_rel_eq_3}
    \alpha^{v}_{ij}= (\Umat \vv_{i}^{\prime})^\top \cdot\Vmat \vv_{j}^{\prime} \,,
\end{align}
where $\Umat, \Vmat \in \R^{d_h \times (d_q+d_v)}$ are projection matrices.
$\alpha^{b}_{ij}$ measures the relative geometric position between any pair of regions:
\begin{align}\label{ref:hid_rel_eq_4}
    \alpha^{b}_{ij}= \max\{0, \wv \cdot f_b(\bv_{i},\bv_{j})\}\,,
\end{align}
where $f_b(\cdot,\cdot)$ first computes a 4-dimensional relative geometry feature $(\log(\frac{|x_i - x_j|}{w_i}),\log(\frac{|y_i - y_j|}{h_i}),\log(\frac{w_j}{w_i}),\log(\frac{h_j}{h_i}))$, then embeds it into a $d_h$-dimensional feature by computing cosine and sine functions of different wavelengths. $\wv \in \R^{d_h}$ transforms the $d_h$-dimensional feature into a scalar weight, which is further trimmed at 0.
Unlike how we assume \texttt{no-relation} for objects that are too far away from each other in the explicit relation setting, the restrictions for implicit relations are learned through $\wv$ and the zero trimming operation. %

\paragraph{Explicit Relation}
We consider semantic relation encoder first. Since edges in the semantic graph $\Ecal_{sem}$ now contain label information and are directional, we design the attention mechanism in (\ref{ref:hid_rel_eq_1}) to be sensitive to both directionality ($v_i$-to-$v_j$, $v_j$-to-$v_i$ and $v_i$-to-$v_i$)
and labels. Specifically, 
\begin{align}
    \vv_{i}^{\star} &= \sigma\Big(\sum_{j \in \mathcal{N}_i} \alpha_{ij}\cdot (\Wmat_{dir(i,j)}\vv_j^{\prime} +\bv_{lab(i,j)} \Big)\,,\\
    \alpha_{ij}&= \frac{\exp((\Umat \vv_{i}^{\prime})^\top \cdot\Vmat_{dir(i,j)} \vv_{j}^{\prime}+ \cv_{lab(i,j)})}{\sum_{j\in \Ncal_i} \exp((\Umat \vv_{i}^{\prime})^\top \cdot\Vmat_{dir(i,j)} \vv_{j}^{\prime}+\cv_{lab(i,j)})} \nonumber \,,
\end{align}
where $\Wmat_{\{\cdot\}}, \Vmat_{\{\cdot\}}$ are matrices, and $\bv_{\{\cdot\}}, \cv_{\{\cdot\}}$ are bias terms. $dir(i,j)$ selects the transformation matrix wrt the directionality of each edge, and $lab(i,j)$ represents the label of each edge.  Consequently, after encoding all the regions $\{\vv^{\prime}_i\}_{i=1}^{K}$ via the above graph attention mechanism, the refined region-level features $\{\vv^{\star}_i\}_{i=1}^{K}$ are endowed with the prior semantic relations between objects. 

As opposed to graph convolutional networks, this graph attention mechanism effectively assigns different weights of importance to nodes of the same neighborhood. Combining with the question-adaptive mechanism, the learned attention weights can reflect which relations are relevant to a specific question. The relation encoder can work in the same manner on the spatial graph $\Ecal_{spa}$, with a different set of parameters to be learned, thus details are omitted for simplicity.   

\subsection{Multimodal Fusion and Answer Prediction}\label{sec:multimodal}
After obtaining relation-aware visual features, we want to fuse question information $\qv$ with each visual representation $\vv_{i}^{\star}$ through a multi-model fusion strategy. Since our relation encoder preserves the dimensionality of visual features, it can be incorporated with any existing multi-modal fusion method to learn a joint representation $\mathbf{J}$:
\begin{align}
    \mathbf{J} &= f(\mathbf{\vv}^{\star},\qv;\Theta) \,,
\end{align}
where $f$ is a multi-modal fusion method and $\Theta$ are trainable parameters of the fusion module.
%

For the Answer Predictor, we adopt a two-layer multi-layer perceptron (MLP) as the classifier, with the joint representation $\mathbf{J}$ as the input. Binary cross entropy is used as the loss function, similar to \cite{anderson2018bottom}. 

In the training stage, different relations encoders are trained independently. In the inference stage, we combine the three graph attention networks with a weighted sum of the predicted answer distributions. Specifically, the final answer distribution is calculated by:
\begin{align*}
    Pr(a = a_i) &= \alpha Pr_{sem}(a = a_i) +\beta Pr_{spa}(a=a_i)  \\
    & +(1-\alpha -\beta)Pr_{imp}(a = a_i)\,, \numberthis \label{eq:trade-off}
\end{align*}
where $\alpha$ and $\beta$ are trade-off hyper-parameters ($0 \leq \alpha + \beta \leq 1, 0\leq \alpha, \beta \leq 1)$. $Pr_{sem}(a = a_i)$, $Pr_{spa}(a=a_i)$ and $Pr_{imp}(a = a_i)$ denote the predicted probability for answer $a_i$, from the model trained with semantic, spatial and implicit relations, respectively.

\section{Experiments}
We evaluate our proposed model on VQA 2.0 and VQA-CP v2 datasets~\cite{VQA,goyal2017making,vqa-cp}. In addition, Visual Genome~\cite{krishnavisualgenome} is used to pre-train the semantic relation classifier. It is also used to augment the VQA dataset when testing on the test-dev and test-std splits. We use accuracy as the evaluation metric:
\begin{align}
    \text{Acc}(\text{ans}) = \min(1, \frac{\text{\#humans provided ans}}{3})\,.
\end{align}

\subsection{Datasets}

\textbf{VQA 2.0} dataset is composed of real images from MSCOCO~\cite{lin2014microsoft} with the same train/validation/test splits. For each image, an average of 3 questions are generated. These questions are divided into 3 categories: \texttt{Y/N, Number} and \texttt{Other}. 10 answers are collected for each image-question pair from human annotators, and the most frequent answer is selected as the correct answer. Both open-ended and multiple-choice question types are included in this dataset. In this work, we focus on the open-ended task, and take the answers that appeared more than 9 times in the training set as candidate answers, which produces $3,129$ answer candidates. The model is trained on the training set, but when testing on the test set, both training and validation set are used for training, and the max-probable answer is selected as the predicted answer.

\textbf{VQA-CP v2} dataset is a derivation of the VQA 2.0 dataset, which was introduced to evaluate
and reduce the question-oriented bias in VQA models. In
particular, the distribution of answers with respect to question types differs between training and test splits.

\textbf{Visual Genome} contains 108K images with densely annotated objects, attributes and relationships, which we used to pre-train the semantic relation classifier in our model. We filtered out those images that also appeared in the VQA validation set, and split the relation data into 88K for training, 8K for validation, and 8K for testing. Furthermore, we selected the top-14 most frequent predicates in the training data, after normalizing the predicates with relationship-alias provided in Visual Genome. The final semantic relation classifier is trained over 14 relation classes plus a \texttt{no-relation} class.

\subsection{Implementation Details} 

Each question is tokenized and each word is embedded using 600-dimensional word embeddings (including 300-dimensional GloVe word embeddings \cite{pennington2014glove}). The sequence of embedded words is then fed into GRU for each time step up to the 14th token (similar to \cite{kim18bilinear}). Questions shorter than 14 words are padded at the end with zero vectors. The dimension of the hidden layer in GRU is set as 1024.
We employ multi-head attention with 16 heads for all three graph attention networks. The dimension of relation features is set to 1024. For implicit relation, we set the embedded relative geometry feature dimension $d_h$ to be 64.  
 
For the semantic relation classifier, we extract pre-trained object detection features with known bounding boxes from Faster R-CNN~\cite{ren2015faster} model in conjunction with ResNet-101~\cite{he2016deep}. More specifically, the features are the output of the Pool5 layer after RoI pooling from Res4b22 feature map \cite{yao2018exploring}. The Faster R-CNN model is trained over 1,600 selected object classes and 400 attribute classes, similar to the bottom-up attention \cite{anderson2018bottom}. 

Our model is implemented based on PyTorch \cite{paszke2017automatic}. In experiments, we use Adamax optimizer for training, with the mini-batch size as 256. For choice of learning rate, we employ the warm-up strategy~\cite{goyal2017accurate}. Specifically, we begin with a learning rate of 0.0005, linearly increasing it at each epoch till it reaches 0.002 at epoch 4. After 15 epochs, the learning rate is decreased by 1/2 for every 2 epochs up to 20 epochs. Every linear mapping is regularized by weight normalization and dropout ($p = 0.2$ except for the classifier with $0.5$).

\subsection{Experimental Results}
\begin{table*}[ht]
\small
\centering
 \begin{adjustbox}{scale=0.93,tabular=l | cc |ccc|ccc| c,center}
  \hline			
 Fusion   & \multicolumn{9}{c}{Model}  \\
 \cline{2-10}
 Method & Baseline & BiLSTM  & Imp & Sem & Spa & Imp+Sem & Imp+Spa  & Sem+Spa & All \\
   \hline
BUTD~\cite{anderson2018bottom}& 63.15 (63.38$^\dagger$)& 61.95 & 64.10 &  64.11 & 64.02 & 64.93 & 64.92 & 64.84 & \textbf{65.30}\\
  \hline
MUTAN~\cite{ben2017mutan}& 58.16 (61.36$^\dagger$)& 61.22 & 62.45 & 62.60 & 62.01 & 63.99 & 63.70 & 63.89 & \textbf{64.37}\\
  \hline
BAN~\cite{kim18bilinear} & 65.36\small{$\pm 0.14$}  (65.51$^\dagger$)& 64.55 &65.93 \small{$\pm 0.06$} & 65.97 \small{$\pm 0.05$} & 66.02 \small{$\pm 0.12$} & 66.81 & 66.76 & 66.85 & \textbf{67.18}\\ %
  \hline  
\end{adjustbox}
\vspace{1mm}
\caption{\small{Performance on VQA 2.0 validation set with different fusion methods. Consistent improvements are observed across 3 popular fusion methods, which demonstrates that our model is compatible to generic VQA frameworks. ($\dagger$) Results based on our re-implementations.}}
\vspace{-4mm}
\label{tab:diff_fusion}
\end{table*}
This sub-section provides experimental results on the VQA 2.0 and VQA-CP v2 datasets. By way of design, the relation encoder can be composed into different VQA architectures as a plug-and-play component. In our experiments, we consider three popular VQA models with different multimodal fusion methods: Bottom-up Top-down~\cite{anderson2018bottom} (BUTD), Multimodal Tucker Fusion~\cite{ben2017mutan} (MUTAN), and Bilinear Attention Network~\cite{kim18bilinear} (BAN). Table~\ref{tab:diff_fusion} reports results on the VQA 2.0 validation set in the following setting: 
\begin{itemize}
    \vspace{-2mm}
    \item \textit{Imp} / \textit{Sem} /  \textit{Spa}: only one single type of relation (implicit, semantic or spatial) is used to incorporate bottom-up attention features. 
    \vspace{-2mm}
    \item \textit{Imp+Sem} / \textit{Imp+Spa} /  \textit{Sem+Spa}: two different types of relations are used via weighted sum. 
    \vspace{-2mm}
    \item
    \textit{All}: all three types of relations are utilized, through weighted sum (e.g.: $\alpha = 0.4, \beta = 0.3$). See Eqn.~(\ref{eq:trade-off}) for details.
    \vspace{-2mm}
\end{itemize}

Compared to the baseline models, we can observe consistent performance gain for all three architectures after adding the proposed relation encoder. These results demonstrate that our ReGAT model is a generic approach that can be used to improve state-of-the-art VQA models. Furthermore, the results indicate that each single relation helps improve the performance, and pairwise combination of relations can achieve consistent performance gain. When all three types are combined, our model can achieve the best performance. Our best results are achieved by combining the best single relation models through weighted sum.  To verify the performance gain is significant, we performed t-test on the results of our BAN baseline and our proposed model with each single relation. We report the standard deviation in Table~\ref{tab:diff_fusion}, and the p-value is $0.001459$. The improvement from our method is significant at $p < 0.05$. We also compare with an additional baseline model that uses BiLSTM as the contextualized relation encoder, the results show that using BiLSTM hurts the performance.

\begin{table}[t!]
\centering
\small
\begin{adjustbox}{scale=0.94,tabular=c|cccccc,center}
  \hline			
  Model &  SOTA \cite{cadene2019murel}  & Baseline & Sem & Spa & Imp & All \\
  \hline
  Acc. & 39.54& 39.24 & 39.54 & 40.30 & 39.58 & \textbf{40.42}\\
  \hline
\end{adjustbox}
\vspace{1mm}
\caption{\small{Model accuracy on the VQA-CP v2 benchmark (open-ended setting on the test split).}}
\label{tab:cp-v2}
\vspace{-2mm}
\end{table}

\begin{table}[t!]
\small
  \centering
 \begin{adjustbox}{scale=0.92,tabular=c|cccc|c,center}
   \hline
 \multirow{2}{*}{Model} &\multicolumn{4}{c|}{Test-dev} & \multirow{2}{*}{Test-std}\\
\cline{2-5}
& Overall & Y/N & Num & Other & \\
    \hline
BUTD~\cite{teney2017tips}   & 65.32 & 81.82 & 44.21 & 56.05 & 65.67\\
   MFH~\cite{yu2018beyond}  & 68.76 & 84.27& 50.66& 60.50& - \\
   Counter~\cite{zhang18count} & 68.09 &83.14 &51.62 &58.97 & 68.41\\
    Pythia~\cite{jiang2018pythia}&70.01 & - & - & - & 70.24\\
    BAN~\cite{kim18bilinear}  & 70.04 & 85.42& 54.04 & \textbf{60.52} & 70.35 \\
\hline
v-AGCN ~\cite{yang2018multi} &65.94& 82.39 & \textbf{56.46} &45.93 & 66.17\\
Graph learner~\cite{norcliffe2018learning} & - & - & - & - & 66.18 \\
MuRel~\cite{cadene2019murel}& 68.03 & 84.77 &49.84 &57.85 & 68.41\\
  ReGAT (ours) & \textbf{70.27}& \textbf{86.08} & 54.42& 60.33& \textbf{70.58}\\
  \hline
  \end{adjustbox}
  \vspace{1mm}
  \caption{\small{Model accuracy on the VQA 2.0 benchmark (open-ended setting on the test-dev and test-std split).}}
  \label{tab:v2results}
  \vspace{-4mm}
\end{table}
To demonstrate the generalizability of our ReGAT model, we also conduct experiments on the VQA-CP v2 dataset, where the distributions of the training and test splits are very different from each other. Table~\ref{tab:cp-v2} shows results on VQA-CP v2 test split. Here we use BAN with four glimpses as the baseline model. Consistent with what we have observed on VQA 2.0, our ReGAT model surpasses the baseline by a large margin. With only single relation, our model has already achieved state-of-the-art performance on VQA-CP v2 (40.30 vs. 39.54). When adding all the relations, the performance gain was further lifted to $+0.88$. 

Table~\ref{tab:v2results} shows single-model results on VQA 2.0 test-dev and test-std splits. The top five rows show results from models without relational reasoning, and the bottom four rows are results from models with relational reasoning. Our model surpasses all previous work with or without relational reasoning. Our final model uses bilinear attention with four glimpses as the multimodal fusion method. Compared to BAN~\cite{kim18bilinear}, which uses eight bilinear attention maps, our model outperforms BAN with fewer glimpses. Pythia~\cite{jiang2018pythia} achieved 70.01 by adding additional grid-level features and using 100 object proposals from a fine-tuned Faster R-CNN on the VQA dataset for all images. Our model, without any feature augmentation used in their work, surpasses Pythia's performance by a large margin.

\subsection{Ablation Study}\label{sec:ablation}
\begin{table}[t!]
\small
\centering
 \begin{adjustbox}{scale=0.94,tabular=c c | c c c c,center}
  \hline			
  Att. & Q-adaptive &  Semantic & Spatial & Implicit  & All\\
  \hline
  No & No & 63.20  & 63.04  & n/a & n/a\\
  Yes & No & 63.90 & 63.85  & 63.36 & 64.98\\
  No & Yes & 63.31  &  63.13  & n/a & n/a\\
  Yes & Yes & \textbf{64.11} & \textbf{64.02} & \textbf{64.10} & \textbf{65.30}\\
  \hline  
\end{adjustbox}
\vspace{1mm}
\caption{\small{Performance on VQA 2.0 validation set for ablation study (Q-adaptive: question-adaptive; Att: Attention).}}
\label{tab:ablation}
\vspace{-4mm}
\end{table}

In Table~\ref{tab:ablation}, we compare three ablated instances of ReGAT with its complete form. Specifically, we validate the importance of concatenating question features to each object representation and attention mechanism. All the results reported in Table~\ref{tab:ablation} are based on BUTD model architecture. To remove attention mechanism from our relation encoder, we simply replace graph attention network with graph convolution network, which can also learn node representation from graphs but with simple linear transformation.

Firstly, we validate the effectiveness of using attention mechanism to learn relation-aware visual features. Adding attention mechanism leads to a higher accuracy for all three types of relation. Comparison between line 1 and line 2 shows a gain of $+0.70$ for semantic relation and $+0.81$ for spatial relation. Secondly, we validate the effectiveness of question-adaptive relation features. Between line 1 and line 3, we see a gain of approximately $+0.1$ for both semantic and spatial relations. Finally, attention mechanism and question-adaptive features are added to give the complete ReGAT model. This instance gives the highest accuracy (line 4). Surprisingly, by comparing line 1 and line 4, we can observe that combining graph attention with question-adaptive gives better gain than simply adding the individual gains from the two methods. It is worth mentioning that for implicit relation, adding question-adaptive improves the model performance by $+ 0.74$, which is higher than the gain from question-adaptive for the two explicit relations.  When all the relations are considered, we observe consistent performance gain by adding the question-adaptive mechanism.

\begin{figure}[t!]
	\centering
	\includegraphics[width = 0.46\textwidth]{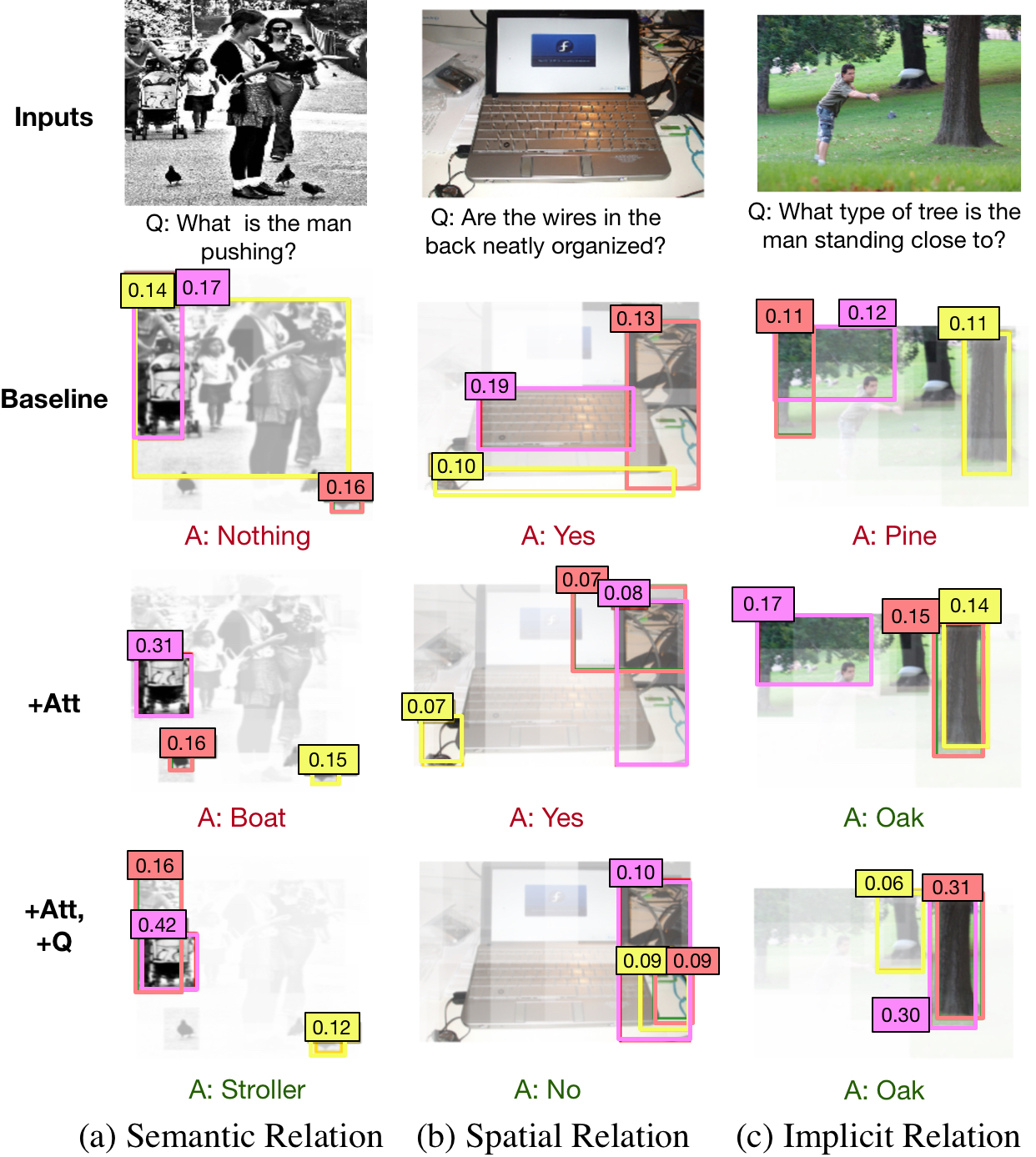}
\caption{\small{Visualization of attention maps learned from ablated instances: The three bounding boxes shown in each image are the top-3 attended regions. The numbers are attention weights.}}
	\label{fig:ablation_vis}
	\vspace{-4mm}
\end{figure}

To better understand how these two components help answer questions, we further visualize and compare the attention maps learned by the ablated instances in Section \ref{sec:vis}. 
\subsection{Visualization}\label{sec:vis}

To better illustrate the effectiveness of adding graph attention and question-adaptive mechanism, we compare the attention maps learned by the complete ReGAT model in a single-relation setting with those learned by two ablated models. As shown in Figure~\ref{fig:ablation_vis}, the second, third and last rows correspond to line 1, 3 and 4 in Table~\ref{tab:ablation}, respectively. Comparing row 2 with row 3 leads to the observation that graph attention helps to capture the interactions between objects, which contributes to a better alignment between image regions and questions. Row 3 and row 4 show that adding the question-adaptive attention mechanism produces sharper attention maps and focuses on more relevant regions. These visualization results are consistent with the quantitative results reported in Table~\ref{tab:ablation}.

\begin{figure}[t]
	\centering
	\includegraphics[width = 0.48\textwidth]{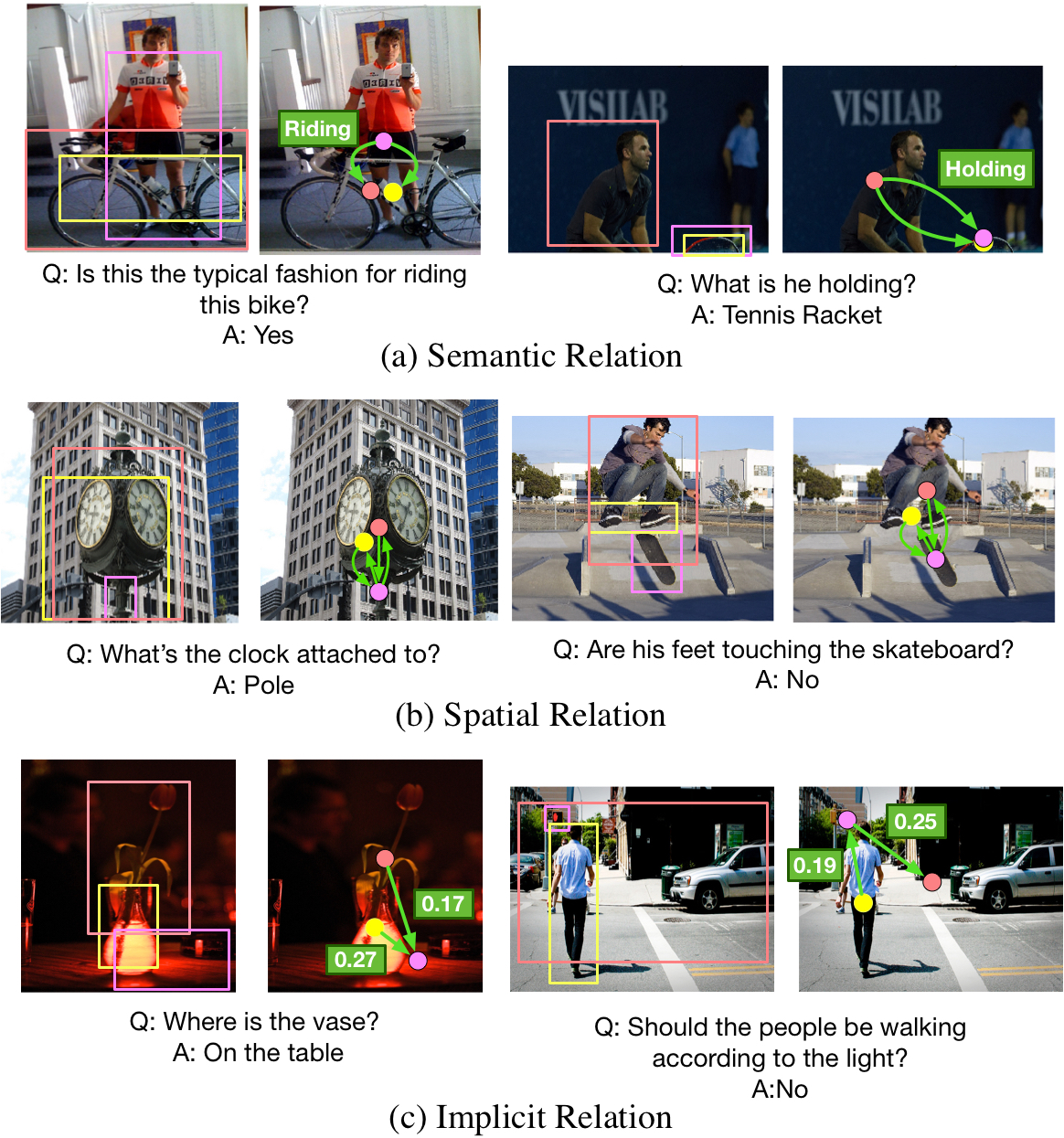}
\caption{\small{Visualization of different types of visual object relation in VQA task. The 3 bounding boxes shown in each image are the top-3 attended regions. Green arrows indicate relations from subject to object. Labels and numbers in green boxes are class labels for explicit relation and attention weights for implicit relation.}}
	\label{fig:example_visual}
	\vspace{-4mm}
\end{figure}

Figure~\ref{fig:example_visual} provides visualization examples on how different types of relations help improve the performance. In each example, we show the top-3 attended regions and the learned relations between these regions. As shown in these examples, each relation type contributes to a better alignment between image regions and questions. For example, in Figure~\ref{fig:example_visual}(a), semantic relations ``\emph{Holding}'' and ``\emph{Riding}'' resonate with the same words that appeared in the corresponding questions. Figure \ref{fig:example_visual}(b) shows how spatial relations capture the relative geometric positions between regions. 

To visualize implicit relations, Figure \ref{fig:example_visual}(c) shows the attention weights to the top-1 region from every other region. Surprisingly, the learned implicit relations are able to capture both spatial and semantic interactions. For example, the top image in Figure \ref{fig:example_visual}(c) illustrates spatial interaction ``\emph{on}'' between the table and the vase, and the bottom image illustrates the semantic interaction ``\emph{walking}'' between the traffic light and the person. 

\section{Conclusion}
We have presented Relation-aware Graph Attention Network (ReGAT), a novel framework for visual question answering, to model multi-type object relations with question-adaptive attention mechanism. ReGAT exploits two types of visual object relations: Explicit Relations and Implicit Relations, to learn a relation-aware region representation through graph attention. Our method achieves state-of-the-art results on both VQA 2.0 and VQA-CP v2 datasets.  The proposed ReGAT model is compatible with generic VQA models. Comprehensive experiments on two VQA datasets show that our model can be infused into state-of-the-art VQA architectures in a plug-and-play fashion.  For future work, we will investigate how to fuse the three relations more effectively and how to utilize each relation to solve specific question types. 

{\small
\bibliographystyle{ieee_fullname}
\bibliography{egbib}
}
\end{document}